%% file: LDSR_arxiv.tex
\documentclass[sigconf,review=false]{acmart}

\usepackage[ruled,linesnumbered]{algorithm2e}
\usepackage{amsmath}
\usepackage{multirow}
\usepackage{booktabs}
\usepackage{microtype}
\usepackage{caption}
\usepackage{subcaption}
\AtBeginDocument{%
  }

\setcopyright{acmlicensed}
\copyrightyear{2024}
\acmYear{2024}
\acmDOI{XXXXXXX.XXXXXXX}

\acmConference[MM '24]{Proceedings of the 32th ACM International Conference on Multimedia}{28 October - 1 November 2024}{Melbourne, Australia
}
\acmISBN{978-1-4503-XXXX-X/18/06}

\begin{document}

\title{Learning Discriminative Spatio-temporal Representations for Semi-supervised Action Recognition}


\author{Yu Wang}
\affiliation{%
  \institution{IAIR, Xi'an Jiaotong University}
  \city{Xi'an, Shaanxi}
  \country{China}
}
\email{wy05140234@stu.xjtu.edu.cn}

\author{Sanping Zhou$^{*}$}
\affiliation{%
  \institution{IAIR, Xi'an Jiaotong University}
  \city{Xi'an, Shaanxi}
  \country{China}}
\email{spzhou@xjtu.edu.cn}

\author{Kun Xia}
\affiliation{%
  \institution{IAIR, Xi'an Jiaotong University}
  \city{Xi'an, Shaanxi}
  \country{China}}
\email{xiakun@stu.xjtu.edu.cn}

\author{Le Wang}
\affiliation{%
  \institution{IAIR, Xi'an Jiaotong University}
  \city{Xi'an, Shaanxi}
  \country{China}}
\email{lewang@xjtu.edu.cn}

\begin{abstract}
    Semi-supervised action recognition aims to improve spatio-temporal reasoning ability with a few labeled data in conjunction with a large amount of unlabeled data.
    Albeit recent advancements, existing powerful methods are still prone to making ambiguous predictions under scarce labeled data, embodied as the limitation of distinguishing different actions with similar spatio-temporal information. 
    In this paper, we approach this problem by empowering the model two aspects of capability, namely discriminative spatial modeling and temporal structure modeling for learning discriminative spatio-temporal representations.
    Specifically, we propose an Adaptive Contrastive Learning~(ACL) strategy. It assesses the confidence of all unlabeled samples by the class prototypes of the labeled data, and adaptively selects positive-negative samples from a pseudo-labeled sample bank to construct contrastive learning.
    Additionally, we introduce a Multi-scale Temporal Learning~(MTL) strategy. It could highlight informative semantics from long-term clips and integrate them into the short-term clip while suppressing noisy information.
    Afterwards, both of these two new techniques are integrated in a unified framework to encourage the model to make accurate predictions. Extensive experiments on UCF101, HMDB51 and Kinetics400 show the superiority of our method over prior state-of-the-art approaches.

\end{abstract}




\maketitle

\section{Introduction}
\begin{figure}[t]
	\centering
	\includegraphics[width=\linewidth]{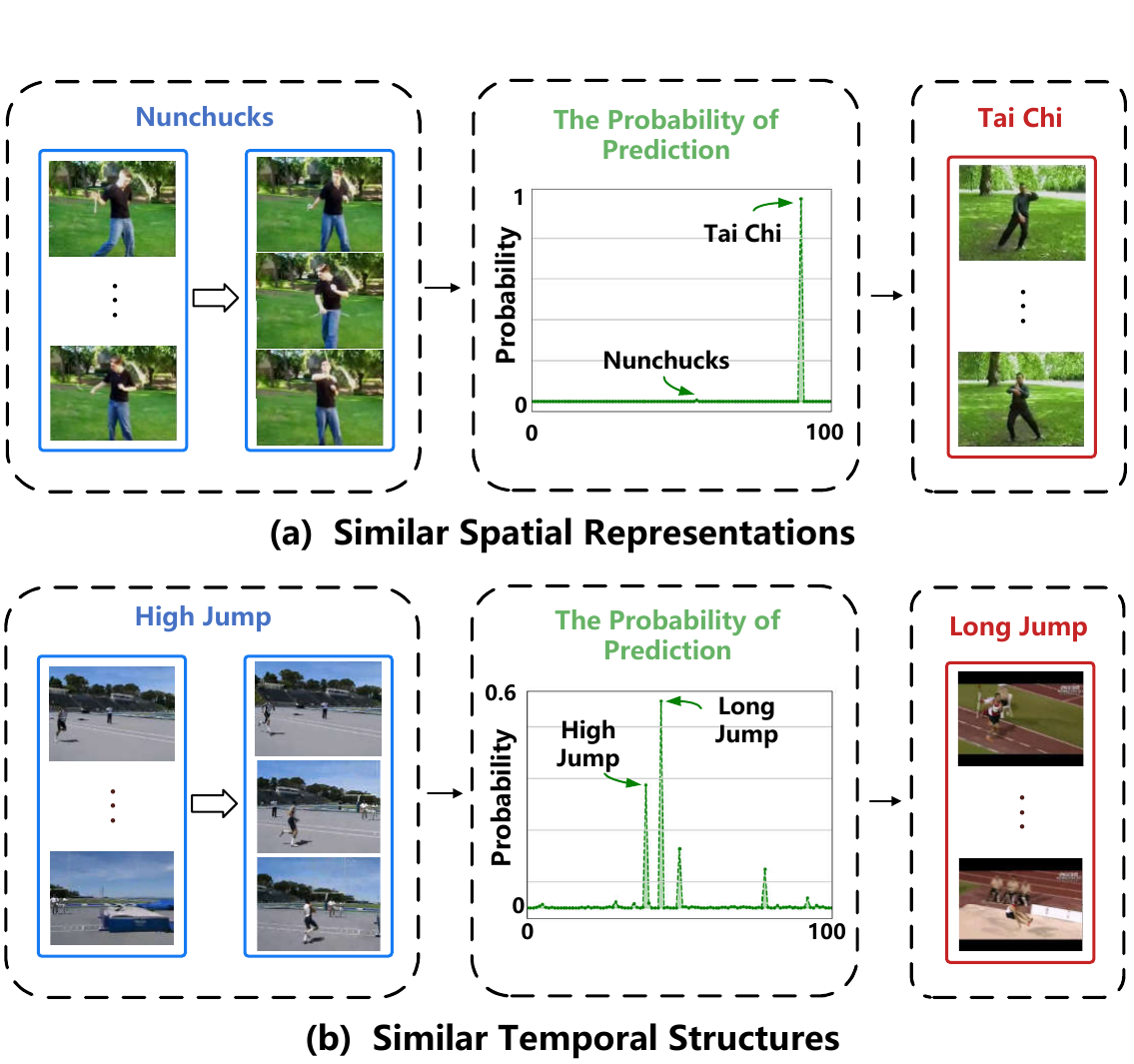}
	\caption{Existing methods are prone to making ambiguous predictions for the actions with similar spatio-temporal semantics. From Figure~\ref{fig1}~(a), the model misrecognizes the action of ``Nunchucks'' as ``Tai Chi'' because of their similar spatial information. From Figure~\ref{fig1}~(b), it is also difficult to enable the model to distinguish between two actions of ``High Jump'' and ``Long Jump'' that have similar sub-actions and temporal structures.}
  \label{fig1}
\end{figure}

Action recognition is one of the most basic topics in video understanding, which has been widely applied to many real-world scenarios, including human-computer interaction~\cite{shu2019hierarchical}, autonomous driving~\cite{wang2018networking}, and so on~\cite{weinland2011survey, poppe2010survey}. However, most existing methods~\cite{zhu2020actbert,wu2021coarse, liu2022video, li2022coarse, bertasius2021space} heavily rely on large-scale and well-annotated training data, which is very tedious and time-consuming.
Therefore, semi-supervised action recognition has attracted growing attentions in academia and industry, which only requires a few labeled data along with a large amount of unlabeled data.
Existing performant methods have achieved remarkable successes in the past few years.
MvPL~\cite{xiong2021multiview} leverages more indicative information from different views by introducing additional temporal gradients and optical flow modalities.
CMPL~\cite{xu2022cross} introduces auxiliary networks to obtain more pseudo labels for efficient semi-supervised learning. 
LTG~\cite{xiao2022learning} improves the representation of the RGB data branch with the help of the temporal gradient modality. 
TimeBalance~\cite{dave2023timebalance} proposes to learn temporally-invariant and temporally-distinctive features to improve the video representation.
In contrast, SVFormer~\cite{xing2023svformer} achieves new state-of-the-art performance arguably due to its powerful capability in modeling inter-frame relationships.

Albeit achieving notable advances, they are still prone to making ambiguous predictions under only scarce labeled data, limiting the model's ability to distinguishing different actions with similar spatio-temporal information.
Two illustrative examples from the powerful method SVFormer~\cite{xing2023svformer} are depicted in Figure~\ref{fig1}. 
An action of ``Nunchucks'' is mistakenly recognized as ``Tai Chi'' due to similar spatial representations and backgrounds between them. Besides, the model is prone to make biased predictions for an action of ``High Jump'' since it shares similar sub-actions and temporal structures with “Long Jump”, hindering the model’s performance.

Building upon the above observations, we propose a new paradigm for semi-supervised action recognition by emphasizing learning discriminative spatio-temporal representations. This paradigm involves an Adaptive Contrastive Learning~(ACL) strategy and a Multi-scale Temporal Learning~(MTL) strategy, both of which encourage the model to make accurate predictions.

Concretely, ACL aims at improving the model's ability of discriminative spatial modeling. 
During training iteration process, it dynamically constructs class prototypes by using labeled samples and updates a momentum memory bank that stores pseudo-labeled samples.
We compute the distance between each pseudo-labeled sample and its corresponding class prototype and encode the distance as its reliability score by a two-component Gaussian Mixture Model~(GMM).
For an unlabeled sample with higher reliability score, we construct contrastive learning by selecting samples of the same category with a reliability score above the threshold from the memory bank as positive samples and the rest as negatives. If it has lower reliability score, we take the whole memory bank as negative samples and its weakly-augmented sample as the positive sample for contrastive learning.
Thus, the proposed ACL empowers the model to learn the discriminative spatial representations between semantic actions.


Additionally, we present a MTL strategy to equip the model with temporal structure modeling capability. Specifically, MTL first obtains long-term clips of different scales via different sampling intervals on unlabeled videos in addition to the original short-term clip. 
Next, we design a cross-scale temporal calibration module, which  could highlight informative semantics from long-term clips and integrate them into the short-term clip while suppressing noisy information. 
Finally, we align the representations from the calibrated long-term clips and the short-term clip, enabling the model to learn the temporal differences between different scales of clips.

The main contributions of this paper can be summarized as follows:
\begin{itemize}
    \item This paper approaches semi-supervised action recognition from a new perspective by learning discriminative spatio-temporal representations. It encourages the model to distinguish different actions with similar spatio-temporal information under scarce labeled data.
    \item We propose an Adaptive Contrastive Learning~(ACL) strategy and a Multi-scale Temporal Learning~(MTL) strategy to empower the model two aspects of capability, \textit{i.e.}, discriminative spatial modeling and temporal structure modeling.
    \item We integrate ACL and MTL into a unified framework, which significantly advances state-of-the-art results. Extensive experiments on the UCF101, HMDB51 and Kinetics400 datasets demonstrate the effectiveness of the proposed framework.
\end{itemize}

\begin{figure*}[t]
    \includegraphics[width=\linewidth]{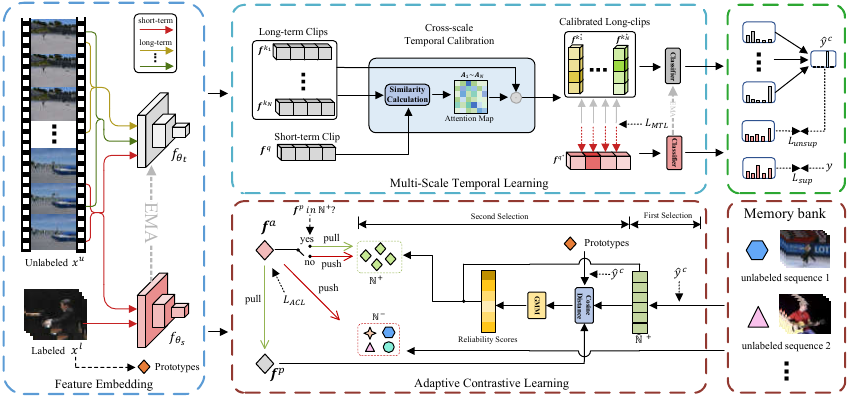}
    \caption{An overview of the proposed Learning Discriminative Spatio-temporal Representations framework. It consists of three parts: (1) a basic framework, including a teacher model for providing pseudo-labels and a student model for online learning, (2) Adaptive Contrastive Learning~(ACL), and~(3) Multi-scale Temporal Learning~(MTL). The labeled portion of the input consists of short-term clips from labeled samples, while the unlabeled portion consists of short-term clips and long-term clips at different scales from unlabeled samples. }
    \label{fig: pipeline}
\end{figure*}

\section{Related Work}
\noindent\textbf{Semi-supervised Learning In Image Classification.}
Recently, semi-supervised learning has been widely explored in the field of image classification. Previous studies predominantly adopt consistency regularization~\cite{xie2020unsupervised,miyato2018virtual,sajjadi2016regularization} or pseudo-labeling~\cite{tarvainen2017mean,lee2013pseudo, arazo2020pseudo, shi2018transductive, rizve2022towards}. Consistency regularization encourages the model to have similar outputs for similar inputs, which requires the model to be robust to perturbations. The pseudo-labeling method requires the model to assign pseudo-labels to unlabeled images based on predictions. Some of the recent
works ~\cite{sohn2020fixmatch,wang2022freematch,zhang2021flexmatch} combine consistency regularization and pseudo-labeling, requiring strong augmented predictions to match weakly augmented ones. In addition, another set of methods ~\cite{zhai2019s4l,assran2021semi} demonstrates the effectiveness of self-supervision in this field. However, these methods do not model the temporal structure of human actions and do not consider discriminative  spatial representations between different actions, resulting in poor performance in the field of semi-supervised action recognition.

\noindent\textbf{Semi-supervised Learning In Action Recognition.}
Recently, significant progress has been made in the field of semi-supervised action recognition. 
MvPL~\cite{xiong2021multiview} and LTG~\cite{xiao2022learning} achieve higher quality pseudo-labels by introducing additional modalities, such as optical flow, temporal gradients, etc. TCL~\cite{singh2021semi} explores the impact of group contrast loss. CMPL~\cite{xu2022cross} introduces an auxiliary network during training, providing complementary gains for unlabeled videos. Timebalance~\cite{dave2023timebalance} learns complementary temporal information through various forms of self-supervised learning. SVFormer~\cite{xing2023svformer} demonstrates the superiority of transformer architecture in the semi-supervised action recognition field and proposes corresponding  data augmentation methods. Compared to these methods, our approach focuses on learning the discriminative spatio-temporal representations, which does not rely on additional modalities and outperforms the state of the art methods on multiple public benchmarks.

\noindent\textbf{Contrastive Learning In Image and Video.}
The essence of contrastive learning lies in maximizing the similarity between positive sample pairs and encouraging discrimination among negative samples. In recent years, significant progress has been made in the field of image contrastive learning~\cite{caron2020unsupervised,chen2020simple,grill2020bootstrap, peng2022crafting,dwibedi2021little, he2020momentum,misra2020self,chen2020big,chen2021empirical,caron2021emerging,song2023semantics}. SimCLR~\cite{chen2020simple} trains a single encoder network with a large batch size to ensure sufficient positive and negative samples for learning. 
BYOL~\cite{grill2020bootstrap} introduces a prediction head concept, employing an asymmetric architecture for contrastive learning without negative samples. 
In addition, notable advancements occurs in video contrastive learning~\cite{dave2022tclr,tao2020self,wang2021enhancing,yao2021seco,wang2022long,behrmann2021long,dorkenwald2022scvrl,ji2024transfer}. 
Compared to image contrastive learning, video contrastive learning also explores enabling models to learn rich temporal variations through contrastive learning. Some methods~\cite{wang2021enhancing,dorkenwald2022scvrl} model the positions of frames, tending to learn inter-frame relationships. 
Other methods~\cite{dave2022tclr,behrmann2021long} focus more on the temporal structure of actions, enabling models to understand the relationship between global actions and local sub-actions. What's different, we propose a new strategy for exploring temporal structure modeling in semi-supervised action recognition. At the same time, we integrate labeled data to assess the confidence of unlabeled samples and adaptively select positive and negative samples for spatial contrastive learning.

\section{Method}
This paper introduces a unified framework for semi-supervised action recognition, emphasizing learning discriminative spatio-temporal representations to enable the model to make accurate predictions. An overview of the proposed framework is illustrated in Figure~\ref{fig: pipeline}. Next, we will elaborate our method step by step.
\subsection{Problem Setting}
Semi-supervised action recognition aims to learn a decent model for accurate action recognition from a smaller labeled dataset $X_{L} = \left\{ \left( x_{i}^{l},y_{i} \right) \right\}_{i = 1}^{N_{l}}$ and a larger unlabeled dataset $X_{U} = \left\{ x_{i}^{u} \right\}_{i = 1}^{N_{u}}$, where $N_{l}$ and $N_{u}$ are the numbers of labeled videos and unlabeled videos.
For a labeled sample $x_{i}^{l}$, its ground truth label is an action category $y_{i}$. The major challenge of semi-supervised action recognition is to improve spatio-temporal reasoning ability especially with scarce labeled data.

\subsection{Basic Framework}
\label{3.2}
We adopt the teacher-student network~\cite{tarvainen2017mean} as our basic framework, in which the teacher is implemented as an Exponential Moving Average~(EMA) of the student.
For training on all labeled data $X_{L}$, the student network is optimized through the standard cross-entropy loss:
\begin{equation}
L_{l} = - \frac{1}{N_{l}}{\sum\limits_{i = 1}^{N_{l}}{y_{i}log\mathcal{F}_{\theta_{s}}\left( A_{w}\left( x_{i}^{l} \right) \right)}},
\end{equation}
where $\mathcal{F}_{\theta_{s}}$ represents the student network and $A_{w}(\cdot)$ represents the weak augmentation. 
For training on unlabeled data $X_{U}$, we first apply different intensities of data augmentation to each unlabeled data $x_{i}^{u}$~\cite{sohn2020fixmatch}. Then, we use the teacher model $\mathcal{F}_{\theta_{t}}$ to generate corresponding pseudo labels ${\hat{y}}_{i}$.
Then, the student model $\mathcal{F}_{\theta_{s}}$ is trained on unlabeled data with pseudo-labels using the cross-entropy loss function:
\begin{equation}
L_{u} = - \frac{1}{N_{u}}{\sum\limits_{i = 1}^{N_{u}}{{\mathbb{I}\left( max\left( 
\mathcal{F}_{\theta_{t}}\left( {A_{w}\left( x_{i}^{u} \right)} \right) \right) > \delta \right)\hat{y}_{i}}log\mathcal{F}_{\theta_{s}}\left( A_{s}\left( x_{i}^{u} \right) \right)}},
\end{equation}
where $A_{s}(\cdot)$ represents the strong augmentation. $\delta$ is a predefined fixed threshold. $\mathbb{I}$ denotes an indicator function, which equals 1 when the maximum category probability exceeds $\delta$, and 0 otherwise. It is used to select high-quality pseudo-labels. So, we define the total loss function as:
\begin{equation}
L_{base} = L_{l} + {\alpha L}_{u},
\end{equation}
where $\alpha$ controls the contribution of the unsupervised cross-entropy loss.

\begin{figure}[t]
	\centering
	\includegraphics[width=1.0\linewidth]{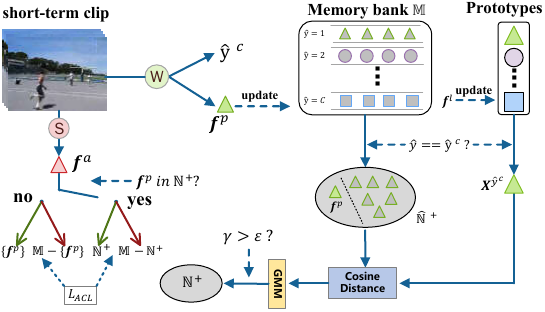}
	\caption{\textbf{Illustration of adaptive contrastive learning module.} We determine the confidence of unlabeled samples and select positive and negative samples for them based on class prototypes.}
	\label{fig: acl.}
\end{figure}

\subsection{Adaptive Contrastive Learning}
\label{3.3}
We present the proposed adaptive contrastive learning strategy in Figure~\ref{fig: acl.}, which aims to enhance the spatial modeling capability of the model. In this module, we first use labeled data $X_{L}$ to construct class prototypes. For $\forall x^{l} \in X_{L}$, we could obtain its feature vector $\textbf{\textit{f}}^{l}$:
\begin{equation}
    \textbf{\textit{f}}^{l} = h_{s}\left( f_{\theta_{s}}\left( A_{w}\left( x^{l} \right) \right) \right).
    \label{(6)}
\end{equation}
where $f_{\theta_{s}}( \cdot )$ and $h_{s}( \cdot )$ represent the encoder and spatial projection head of the student network, respectively. Then, we iteratively update the class prototypes by calculating the exponential moving average of $\textbf{\textit{f}}^{l}$:
\begin{equation}
    \textbf{\textit{X}}_{(t)} = \left( {1 - \beta} \right) \cdot \textbf{\textit{f}}^{l} + \beta \cdot \textbf{\textit{X}}_{(t-1)},
    \label{(7)}
\end{equation}
where $\textbf{\textit{X}}_{(t)}$ and $\textbf{\textit{X}}_{(t-1)}$ are the prototype of an arbitrary class at the $t$-th iteration and $(t-1)$-th iteration respectively. $\beta$ is empirically set to 0.9. 

For $\forall x^{u} \in X_{U}$, we first apply strong augmentation $A_{s}(\cdot)$ and weak augmentation $A_{w}(\cdot)$ to it. Then, we feed them separately into the student and the teacher encoder with spatial projection heads to generate the anchor sample $\textbf{\textit{f}}^{a}$ and its naive positive sample $\textbf{\textit{f}}^{p}$:
\begin{equation}
    \left\{ \begin{matrix}
{\textbf{\textit{f}}^{a} = h_{s}\left( f_{\theta_{s}}\left( A_{s}\left( x^{u} \right) \right) \right)} \\
{\textbf{\textit{f}}^{p} = h_{t}\left( f_{\theta_{t}}\left( A_{w}\left( x^{u} \right) \right) \right)}
\end{matrix} \right.,
\label{(4)}
\end{equation}
where $f_{\theta_{t}}( \cdot )$ and $h_{t}( \cdot )$ represent the encoder and spatial projection head of the teacher network respectively. The parameters of $h_{t}( \cdot )$ are the exponential moving average of $h_{s}( \cdot )$. In order to ensure the diversity of spatial features, we establish a momentum memory bank $\mathbb{M}$~\cite{he2020momentum} that dynamically stores each unlabeled sample $\textbf{\textit{f}}$ with its pseudo label $\hat{y}$ from the teacher network during model training. 

Key aspect of efficient contrastive learning lies in selecting high-quality positive and negative samples. An intuitive idea is to choose the samples of the same category from $\mathbb{M}$ as reliable positives. 
Given a training sample $x^{u}$ with its pseudo label ${\hat{y}}^{c}$, we could construct an initial positive sample set ${\hat{\mathbb{N}}}^{+}$:
\begin{equation}
    {\hat{\mathbb{N}}}^{+} = \left\{ \textbf{\textit{f}} \middle| {\hat{y} = {\hat{y}}^{c},\textbf{\textit{f}} \in \mathbb{M}} \right\} \cup \left\{ \textbf{\textit{f}}^{p} \right\}.
\label{(5)}
\end{equation}

\begin{algorithm}[t]
\caption{Adaptive Contrastive Learning}
\label{acl_alg}
		\SetAlgoLined			
		\DontPrintSemicolon		
        \SetKwInOut{Require}{\textbf{Require}}		
        \SetKwInOut{Require}{\textbf{Require}}	
		\SetKwInOut{Require}{\textbf{Require}}		
        \SetKwInOut{Require}{\textbf{Require}}
        
		\Require{Labeled dataset $X_{L}$, unlabeled dataset $X_{U}$}
		\Require{Momentum memory bank $\mathbb{M}$, threshold $\varepsilon$}
        \Require{Teacher encoder $f_{\theta_{t}}$, student encoder $f_{\theta_{s}}$, spatial projection head $h_{t}( \cdot )$ and $h_{s}( \cdot )$}
		\For{$x^{l} \in X_{L}, {x}^{u} \in X_{U}$}    
			{   
                Update prototype $\textbf{\textit{X}}_{(t)}$ with Eq.~\ref{(6)} and Eq.~\ref{(7)};
                
				Get anchor $\textbf{\textit{f}}^{a}$ and naive positive sample $\textbf{\textit{f}}^{p}$ with Eq.~\ref{(4)};

                Initially select positive sample set ${\hat{\mathbb{N}}}^{+}$ from $\mathbb{M}$ with Eq.~\ref{(5)};

                Compute cosine distance $dis\left( {\textbf{\textit{f}}_{i},\textbf{\textit{X}}^{{\hat{y}}^{c}}} \right)$ with Eq.~\ref{(8)};

                Obtain reliability scores $\gamma_{i}$ according to Eq.~\ref{(9)};

                Combine $\varepsilon$ to further select $\mathbb{N}^{+}$ according to Eq.~\ref{final_sec};
                
                \If{$\textbf{\textit{f}}^{p} \in \mathbb{N}^{+}$}
		     {
			        Calculate $L_{ACL}$ according to Eq.~\ref{(10)};
		     }
             \Else
		    {
			        $\mathbb{N}^{+} = \left\{ \textbf{\textit{f}}^{p} \right\}$, calculate $L_{ACL}$ according to Eq.~\ref{(10)};
	     	}
			}

\end{algorithm}
However, pseudo labels may be noisy and unreliable. Thus, we further evaluate their confidences based on the class prototypes. 
We first compute the cosine distances between the samples from ${\hat{\mathbb{N}}}^{+}$ and the corresponding class prototypes.
For the $i$-th sample $\textbf{\textit{f}}_{i}$ in ${\hat{\mathbb{N}}}^{+}$, the cosine distance could be formulated as:
\begin{equation}
dis\left( {\textbf{\textit{f}}_{i},\textbf{\textit{X}}^{{\hat{y}}^{c}}} \right) = \frac{\textbf{\textit{f}}_{i} \cdot \textbf{\textit{X}}^{{\hat{y}}^{c}}}{{\parallel \textbf{\textit{f}}_{i} \parallel}_{2} \cdot {\parallel \textbf{\textit{X}}^{{\hat{y}}^{c}} \parallel}_{2}},
\label{(8)}
\end{equation}
where ${\parallel \cdot \parallel}_{2}$ is the $L_{2}$ norm. Afterwards, we input all $dis\left( {\textbf{\textit{f}}_{i},\textbf{\textit{X}}^{{\hat{y}}^{c}}} \right)$ into a two-component Gaussian Mixture Model (GMM) to generate the reliability score $\gamma_{i}$ for each $\textbf{\textit{f}}_{i}$:
\begin{equation}
    \gamma_{i} = GMM\left( dis\left( {\textbf{\textit{f}}_{i},\textbf{\textit{X}}^{{\hat{y}}^{c}}} \right) \middle| \left\{ dis\left( {\textbf{\textit{f}}_{i},\textbf{\textit{X}}^{{\hat{y}}^{c}}} \right) \right\}_{i = 1}^{\left| {\hat{\mathbb{N}}}^{+} \right|} \right),
\label{(9)}
\end{equation}
where $\gamma_{i} \in \lbrack 0,1\rbrack$. To select reliable positive samples from ${\hat{\mathbb{N}}}^{+}$ based on reliability scores, we set a threshold $\varepsilon$ and consider those with reliability scores greater than $\varepsilon$ as the final positive samples, denoted as 
\begin{equation}
    \mathbb{N}^{+} = \left\{ \textbf{\textit{f}}_{i} \middle| \gamma_{i} > \varepsilon,{\textbf{\textit{f}}}_{i} \in {\hat{\mathbb{N}}}^{+} \right\}.
\label{final_sec}
\end{equation}

Therefore, we could construct contrastive learning for $x^{u}$ by selecting positive samples from $\mathbb{N}^{+}$ and negative samples from $\mathbb{M}-\mathbb{N}^{+}$, respectively. However, the reliability score of $x^{u}$ may be very low due to the limited performance of the model under scarce labeled data, resulting in its pseudo label error. So, we construct contrastive learning through taking the whole memory bank $\mathbb{M}$ as the negative samples and only $\textbf{\textit{f}}^{p}$ as the positive sample when its $\gamma \leq \varepsilon$.

Finally, the loss of adaptive contrastive learning is expressed in the following formula:
\begin{equation}
L_{ACL} = - log\frac{\left. {\sum_{\textbf{\textit{f}} \in \mathbb{N}^{+}}{exp\left( \textbf{\textit{f}} \right.}^{a}} \cdot \textbf{\textit{f}}/\tau \right)}{\left. {\sum_{\textbf{\textit{f}} \in \mathbb{N}^{+}}{exp\left( \textbf{\textit{f}} \right.}^{a}} \cdot \textbf{\textit{f}}/\tau \right)\left. + {\sum_{\textbf{\textit{f}} \in \mathbb{M}-\mathbb{N}^{+}}{exp\left( \textbf{\textit{f}} \right.}^{a}} \cdot \textbf{\textit{f}}/\tau \right)}
\label{(10)}
\end{equation}
where $\tau$ is the temperature hyperparameter. During the training process, we select different positive and negative samples based on the reliability score of $x_{u}$. The summary of ACL is presented in Algorithm~\ref{acl_alg}.

\subsection{Multi-scale Temporal Learning}
\label{3.4}
\begin{figure}[t]
	\centering
	\includegraphics[width=1.0\linewidth]{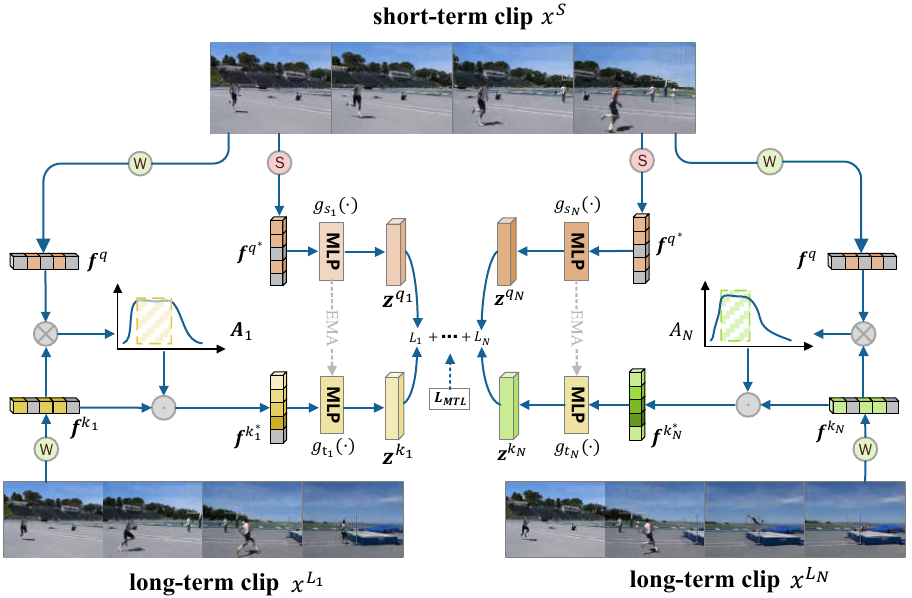}
	\caption{\textbf{Illustration of multi-scale temporal learning module.} We calibrate long-term clips of different scales and align them with the short-term clip.}
	\label{fig: cross attention.}
\end{figure}

As in the previous analysis, the model is prone to make ambiguous predictions for different actions that share similar temporal structures, especially when few labeled data is available. To this end, we propose a Multi-scale Temporal Learning~(MTL) strategy to improve the ability of model's temporal structure modeling, as shown in Figure~\ref{fig: cross attention.}.

Given an unlabeled video $x^u \in X_U$, we randomly sample a short-term clip and multiple long-term clips by using different sampling rates. All clips include $T$ frames and we represent the short-term clip as $x^S$ and the long-term clips as $x^{L_n}$, where $n = 1, 2, \ldots, N$. $N$ is a hyperparameter that controls the number of long-term clips and $N=2$ in the experiments.
Afterwards, we enter them into the teacher network and the student network respectively to obtain the query and key:
\begin{equation}
    \left\{ \begin{matrix}
{\textbf{\textit{f}}^{q} = ~f_{\theta_{s}}\left( A_{w}\left( x^{S} \right) \right)} \\
{\textbf{\textit{f}}^{k_n} = ~f_{\theta_{t}}\left( A_{w}\left( x^{L_n} \right) \right)}
\end{matrix} \right..
\end{equation}
Then, we design a cross-scale temporal calibration module, aiming to suppress the parts in each scale of long-term clips that are unrelated to the short-term clip. We calculate the similarity between $\textbf{\textit{f}}^{q}$ and $\textbf{\textit{f}}^{k_{n}}$ as follows:
\begin{equation}
    \textbf{\textit{A}}_{n} = \frac{\textbf{\textit{f}}^{q} \cdot \textbf{\textit{f}}^{k_n}}{{\parallel \textbf{\textit{f}}^{q} \parallel}_{2} \cdot {\parallel \textbf{\textit{f}}^{k_n} \parallel}_{2}},
\end{equation}
where $\parallel \cdot \parallel$ denotes the $L_{2}$-norm. $\textbf{\textit{A}}_{n}$ is the generated attention map, where semantic positions unrelated to the short-term clip are assigned low weights. Then, we calibrate long-term clips by multiplying $\textbf{\textit{A}}_{n}$ with $\textbf{\textit{f}}^{k_n}$:
\begin{equation}
    \textbf{\textit{f}}^{k^{*}_n} = \textbf{\textit{A}}_{n} \cdot \textbf{\textit{f}}^{k_n},
\end{equation}
where $\textbf{\textit{f}}^{k^{*}_n}$ represents the representation of the calibrated long-term clip. We also apply strong enhancement to the short-term clip and encode it as feature $\textbf{\textit{f}}^{q^{*}}$:
\begin{equation}
    \textbf{\textit{f}}^{q^{*}} = f_{\theta_{s}}\left( A_{s}\left( x^{S} \right) \right).
\end{equation}

Then, we feed both of them to their respective temporal projection heads to obtain the final representations:
\begin{equation}
    \left\{ \begin{matrix}
{\textbf{\textit{z}}^{q_{n}} = g_{s_{n}}\left( \textbf{\textit{f}}^{q^{*}} \right)} \\
{\textbf{\textit{z}}^{k_{n}} = g_{t_{n}}\left( \textbf{\textit{f}}^{k_{n}^{*}} \right)}
\end{matrix} \right.,
\end{equation}
where $g_{s_{n}}( \cdot )$ and $g_{t_{n}}( \cdot )$ respectively represent the temporal projection heads of the student network and the teacher network.
The parameters of $g_{t_{n}}( \cdot )$ are also the exponential moving average of $g_{s_{n}}( \cdot )$. 
Then, following DINO~\cite{caron2021emerging}, we use softmax function with temperature hyperparameters $\tau_{s}$ and $\tau_{t}$ to normalize the outputs $\textbf{\textit{z}}^{q_{n}}$ and $\textbf{\textit{z}}^{k_{n}}$. The output probability distributions $P^{q_{n}}$ and $P^{k_{n}}$ are used to calculate alignment loss as follows:
\begin{equation}
    L_{n} = H\left( P^{k_{n}},P^{q_{n}} \right),
\end{equation}
where $H\left( {a,b} \right) = - alogb$. The final MTL loss is as follows:
\begin{equation}
    L_{MTL} = \frac{1}{N}{{\sum\limits_{n = 1}^{N}L_{n}}}.
\end{equation}

\subsection{Training Objective}
\label{3.5}
During the training stage, we make some adjustments to ${L}_{base}$ in Section~\ref{3.2}. Firstly, we obtain the pseudo-label for each unlabeled data $x_{i}^{u}$ by taking the average of the predictions of multiple clips in Section~\ref{3.4}:
\begin{equation}
    \begin{aligned}
        {\hat{y}}_{i} = argmax\left( \mathcal{F}_{\theta_{t}}\left( {A_{w}\left( x^{S} \right)} \right) + {\sum\limits_{n = 1}^{N}{\mathcal{F}_{\theta_{t}}\left( A_{w}\left( x^{L_n} \right) \right)}} \right).
    \end{aligned}
\end{equation}
Then, we use the reliability score $\gamma_{i}$ obtained in Section~\ref{3.3} to replace the fixed weight $\alpha$. Finally, the complete loss function is as follows:
\begin{equation}
    \begin{aligned}
        L = L_{base} + \mu_{1} \cdot L_{MTL} + \mu_{2} \cdot L_{ACL},
    \end{aligned}
\end{equation}
where $\mu_{1}$ and $\mu_{2}$ are responsible for controlling the contributions of $L_{MTL}$ and $L_{ACL}$.

\input{Figure/table1}

\section{Experiment}

In this section, we introduce the experimental settings in Section~\ref{4.1} and Section~\ref{4.2}. From Section~\ref{4.3} to Section~\ref{4.4}, we conduct experiments under different labeling rates with ablation experiments and empirical analysis. We also show the visualization in Section~\ref{4.5}. Note that we only use the RGB modality and the official validation set for inference.

\subsection{Datasets and Evaluation}
\textbf{Datasets.}
\label{4.1}
Following the previous performant semi-supervised action recognition methods~\cite{dave2023timebalance, xing2023svformer}, we evaluate our method on three public action recognition benchmarks: UCF-101~\cite{soomro2012ucf101}, HMDB-51~\cite{kuehne2011hmdb}, and Kinetics400~\cite{carreira2017quo}. 

\noindent UCF-101 is a widely used video dataset, which includes 13,220 videos belonging to 101 classes.
We randomly select 1 or 10 samples from each class as labeled sets following SVFormer~\cite{xing2023svformer}. 

\noindent HMDB-51 is a smaller dataset with 6,776 videos across 51 classes. We follow the splits of VideoSSL~\cite{jing2021videossl} and LTG~\cite{xiao2022learning} and conduct experiments at three different labeling rates: 40$\%$, 50$\%$, 60$\%$. 

\noindent Kinetics-400 is a large-scale dataset consisting of approximately 245k training videos and 20k validation videos belonging to 400 classes. For Kinetics-400, we follow SVFormer~\cite{xing2023svformer} and TimeBalance~\cite{dave2023timebalance}, randomly sampling 6 and 60 videos for each class under 1$\%$ and 10$\%$ settings, forming two balanced labeled subsets.

\noindent\textbf{Evaluation Metric.}
We report Top-1 accuracy for major comparisons and Top-5 accuracy in some ablation studies.

\subsection{Implementation Details}
\label{4.2}
\textbf{Baseline.}
We follow SVFormer~\cite{xing2023svformer} combined with TimeSformer~\cite{bertasius2021space} as our baseline, which is also initialized with weights from ImageNet~\cite{deng2009imagenet}. 
Compared with the traditional convolutional neural network~(CNN) based methods, SVFormer could achieve significant performance gains by only requiring the same number of learnable parameters.

\noindent\textbf{Training Details.}
For the training process, we follow the settings of TimeSformer~\cite{bertasius2021space}. All experiments are done with the SGD optimizer during training, with a momentum of 0.9 and weight decay of 0.001. We set the initial learning rate as 0.005 and divide it by 10 at the 25-$th$ epoch and the 28-$th$ epoch. 
The weak augmentation is implemented through random scaling and random cropping while the strong augmentation works by adding additional grayscale and color jitter to the weakly-augmented samples.
In addition, we set the confidence threshold $\delta$ to 0.3 and set temperature hyperparameters$~ \tau$, $\tau_{s}$, and $\tau_{t}$ are set to 0.07, 0.1, and 0.04 respectively. The loss weights $\mu_{1}$ and $\mu_{1}$ are both set to 1. 
According to~\cite{xing2023svformer}, $B_{l}$ and $B_{u}$ are set to 1 and 5 respectively. Each clip consists of 8 frames. $\varepsilon$ is set to 0.7, and it will be further analyzed in subsequent ablation experiments. The sampling steps for short-term clips are 8, and 16 and 32 for long-term clips.

\noindent\textbf{Inference.} During the testing phase, we follow the recent state-of-the-art method~\cite{xing2023svformer}, sampling five segments from the entire video and create three different crops to achieve a resolution of 224×224. The final prediction is the average of the softmax probabilities of these 5×3 predictions.

\subsection{Comparison with State-of-the-art Methods}
\label{4.3}
In order to demonstrate the capability of our proposed method, we compare it with the latest semi-supervised action recognition methods on three public datasets, including UCF101~\cite{soomro2012ucf101}, HMDB51~\cite{kuehne2011hmdb}, and Kinetics400~\cite{carreira2017quo}. 

\noindent\textbf{Comparison with CNN-based methods.}
We report the comparison results with 13 CNN-based methods in Table~\ref{tab:comparison with sota}. In general, our method does not use additional modalities and significantly outperforms all CNN-based methods by a large margin on all labeling ratios.

\input{Figure/abla_components}
\noindent\textbf{Comparison with Transformer-based methods.}
Compared with the transformer-based method~\cite{xing2023svformer}, our proposed method records the new state-of-the-art performance on the three benchmarks. Specifically, our method outperforms it by an average of 7.9$\%$ on UCF101 dataset and 2.1$\%$ on HMDB51 dataset at the same inference cost.
This further confirms the importance and superiority of enhancing  discriminative spatial modeling and temporal structure modeling.



\subsection{Ablation Study}
\label{4.4}
To better understand how the proposed method works, we conduct a series of ablation studies on UCF-101 at the 1$\%$ labeling ratio setting and HMDB51 at the 40$\%$ labeling ratio setting. Our method follows the same semi-supervised learning pattern as FixMatch~\cite{sohn2020fixmatch} and MeanTeacher~\cite{tarvainen2017mean}.


\noindent\textbf{Contributions of different training components.}
To validate our key designs, we ablate the effects of the proposed adaptive contrastive learning~(ACL) and multi-scale temporal learning~(MTL). 
For this purpose, we established four experimental configurations, as shown in Table~\ref{tab:ablation-on-components}: (1) baseline, which corresponds to the basic framework mentioned in Section~\ref{3.2} and has been adjusted according to Section~\ref{3.5} to validate the effect of enhancing only discriminative spatio-temporal representations. (2) only ACL, (3) only MTL, (4) both. It can be observed that the baseline has achieved decent results but still significantly lower than the other three configurations. This indicates that relying solely on pseudo-labels for semi-supervised action recognition is insufficient with limited labeled data. Furthermore, applying our ACL strategy significantly improves performance on both datasets, which suggests that discriminative spatial features play a crucial role for more accurate predictions. Additionally, equipping the baseline model with the MTL strategy alone also lead to significant performance gains, indicating that modeling temporal structures contributes to enhancing the model's discriminative capacity. Finally, the both strategies improve the spatio-temporal reasoning ability of the model in a cooperative way, as achieving both on top of the baseline outperforms either one alone.

\input{Figure/abl_acl1}

\input{Figure/abl_acl2}

\input{Figure/abl_mtl}

\noindent\textbf{Analysis of Adaptive Contrastive Learning.} 
Regarding our ACL module, an intuitive question is how the performance would be if we do not consider the confidence of features. Therefore, we compare the proposed ACL with two alternative contrastive learning based methods, MoCo~\cite{he2020momentum} and SimCLR~\cite{chen2020simple}. 
We conduct an ablation study on the UCF-101 dataset with 1$\%$ labeling ratio and use same data augmentations and experiment setting for fair comparison.
As shown in Table~\ref{tab:acl1}, it can be observed that all methods improve the performance, proving the effectiveness of spatial representation modeling. In addition, we observe that ACL is better than MoCo and SimCLR. It is reasonable because ACL considers sample confidence and reduces noise in contrastive learning by adaptively selecting positive and negative samples, which demonstrates the necessity of considering feature confidence in spatial contrastive learning for semi-supervised action recognition.

\noindent\textbf{Analysis of threshold $\varepsilon$.} 
The threshold $\varepsilon$ is used to evaluate the reliability of samples in ACL. The larger value of $\varepsilon$ means that the fewer samples are assigned as positive samples in the momentum memory bank while the lower $\varepsilon$ may take some unreliable samples with noisy pseudo labels as positive samples, degenerating the model performance. 
To this end, we conduct an ablation study on the choice of hyper-parameter $\varepsilon$. From Table~\ref{tab:acl2}, we can observe that the performance peak around $\varepsilon = 0.7$. In addition, we ablate the case where the confidence of the unlabeled sample $x^{u}$ is not considered. We can see that the performance will decrease due to the introduction of noise.

\noindent\textbf{Analysis of Multi-scale Temporal Learning.}
Learning from multi-scale temporal information could enable the model to learn the temporal structures of different semantic actions.
We conduct an experiment on how different scales of temporal information impacts the model performance. 
As shown in Table~\ref{mtl}, we set up four experimental settings: (1) none, (2) only long-term $x^{L_{1}}$ (3) only long-term $x^{L_{2}}$ and (4) multi-scale temporal information. The experimental results indicate that temporal information at different scales contributes to learning discriminative temporal structures, 
which improves the temporal modeling ability of the model to distinguish different actions with similar sub-actions.

\subsection{Visualization}
\label{4.5}
In Figure~\ref{fig:tsne feature}, we show the visualization of embeddings with and without enhancing discriminative spatio-temporal representation. Generally, due to limited labeled samples in semi-supervised learning, it is difficult to for models to form accurate clusters. We observe that after applying our spatio-temporal representation enhancement, the embeddings of the same categories cluster better together. In addition, we also show the model's predictions for the same blurred  sample, as shown in Figure~\ref{fig:pred}. It can be observed that, after enhancing discriminative spatio-temporal representations, the model exhibits greater discriminability for actions with similar spatio-temporal information.

\input{Figure/tsne}

\subsection{Performance gains}
In Table~\ref{tab:performance}, we observe that our framework achieves an average gain of more than 30$\%$ for categories prone to ambiguous predictions \textit{Javelin Throw}, \textit{Soccer Juggling}, \textit{High Jump}, \textit{Juggling Balls} and \textit{YoYo}. ACL enhances the model's discriminative spatial modeling capabilities and more accurately predicts actions \textit{YoYo} and \textit{Juggling Balls} that rely on spatial representation. For actions like \textit{Javelin Throw} and \textit{High Jump} that rely on temporal context and temporal structure, MTL enables the model to learn temporal information at different scales to enhance the ability to distinguish these actions.

\input{Figure/pred}

\input{Figure/gain}


\section{Conclusion}
In this paper, we propose a new semi-supervised action recognition learning framework, which achieves more accurate predictions by enhancing discriminative spatio-temporal representations. Our method introduces adaptive contrastive learning and multi-scale temporal learning strategies, and integrates them into a unified framework. Our approach significantly surpasses all previous methods and achieves state-of-the-art performance on the UCF-101, HMDB-51, and Kinetics-400 datasets with different label ratios. In the future, we plan to investigate the effectiveness of learning discriminative representations in other video-based tasks and explore new methods for learning discriminative representations.

\newpage
{\small
\bibliographystyle{ACM-Reference-Format}
\bibliography{reference}
}

\end{document}

%% file: Figure/table1.tex
\begin{table*}[t]
\centering
\caption{\textbf{Compare with state-of-the-art methods.} The results are reported with Top-1 accuracy (\%) on the validation sets. V-Video(RGB), F-Optical Flow, G-Temporal Gradients. The best performance of each setting is highlighted in bold.}
\resizebox{1.0\linewidth}{!}{
\begin{tabular}{lcccc|cc|ccc|cc}
\toprule
\multicolumn{1}{c}{\multirow{2}{*}{Method}} & \multirow{2}{*}{Backbone}        & \multirow{2}{*}{Input} & \multirow{2}{*}{w ImgNet} & \multirow{2}{*}{\# F} & \multicolumn{2}{c|}{UCF101} & \multicolumn{3}{c|}{HMDB51} & \multicolumn{2}{l}{Kinetics400} \\ \cline{6-12} 
\multicolumn{1}{c}{}                        &                                  &                        &                           &                       & 1\%          & 10\%         & 40\%    & 50\%    & 60\%    & 1\%            & 10\%           \\ \hline
MT(NeurIPS 2017)~\cite{tarvainen2017mean}                      & 3D-ResNet-18                     & V                      &                           & 16                    & --           & 25.6         & 27.2    & 30.4    & 32.2    & --             & --             \\
SD(ICCV 2019)~\cite{girdhar2019distinit}                        & 3D-ResNet-18                     & V                      &                           & 16                    & --           & 40.7         & 32.6    & 35.1    & 36.3    & --             & --             \\
FixMatch(NeurIPS 2020)~\cite{sohn2020fixmatch}                & SlowFast-R50                     & V                      & $\checkmark$              & 8                     & 16.1         & 55.1         & --      & --      & --      & 10.1           & 49.4           \\
MT+SD(WACV 2021)~\cite{jing2021videossl}                     & \multicolumn{1}{l}{3D-ResNet-18} & V                      &                           & 16                    & --           & 40.5         & 32.3    & 33.6    & 35.7    & --             & --             \\
VideoSSL(WACV 2021)~\cite{jing2021videossl}                    & 3D-ResNet-18                     & V                      & $\checkmark$              & 16                    & --           & 42.0         & 32.7    & 36.2    & 37.0    & --             & 33.8           \\
TCL(CVPR 2021)~\cite{singh2021semi}                        & TSM-ResNet-18                    & V                      &                           & 8                     & --           & --           & --      & --      & --      & 11.6           & --             \\
ActorCM(CVIU 2021)~\cite{zou2023learning}                    & R(2+1)D-34                       & V                      & $\checkmark$              & 8                     & --           & 53.0         & 35.7    & 39.5    & 40.8    & 9.02           & --             \\
MvPL(ICCV 2021)~\cite{xiong2021multiview}                      & 3D-ResNet-50                      & V+F+G                    & \multicolumn{1}{l}{}      & 8                     & 22.8         & 80.5         & --      & --      & --      & 17.0           & 58.2           \\
CMPL(CVPR 2022)~\cite{xu2022cross}                       & R50+R50-1/4                      & V                      & $\checkmark$              & 8                     & 25.1         & 79.1         & --      & --      & --      & 17.6           & 58.4           \\
LTG(CVPR 2022)~\cite{xiao2022learning}                        & 3D-ResNet-18                     & V+G                    &                           & 8                     & --           & 62.4         & 46.5    & 48.4    & 49.7    & 9.8            & 43.8           \\
TACL(TCSVT 2022)~\cite{tong2022semi}                      & 3D-ResNet-18                     & V                      & $\checkmark$              & 16                    & --           & 55.6         & 38.7    & 40.2    & 41.7    & --             & --             \\
L2A(ECCV 2022)~\cite{gowda2022learn2augment}                        & 3D-ResNet-18                     & V                      & $\checkmark$              &  8                     & --           & 60.1         & 42.1    & 46.3    & 47.1    & --             & --             \\
TimeBalance(CVPR 2023)~\cite{dave2023timebalance}                & \multicolumn{1}{l}{3D-ResNet-50} & V                      &                           & 8                     & 30.1         & 81.1         & 52.6    & 53.9    & 54.5    & 19.6           & 61.2           \\
SVFormer(CVPR 2023)~\cite{xing2023svformer}                   & ViT-B                            & V                      & $\checkmark$              & 8                     & 46.3         & 86.7         & 61.6    & 64.4    & 68.2    & 49.1           & 69.4           \\
Ours                                   & ViT-B                            & V                      & $\checkmark$              & 8                     & \textbf{60.1}         & \textbf{88.6}         & \textbf{64.9}    & \textbf{66.9}    & \textbf{68.8}    & \textbf{50.1}           &  \textbf{69.9}              \\ \bottomrule 
\end{tabular}
}
	\label{tab:comparison with sota}
\end{table*}

%% file: Figure/abla_components.tex
\begin{table}[t]
\centering
\caption{
		~\textbf{Contribution of different training components.} Bold indicates the best results. Results are reported on UCF-101 with 1$\%$ labeled setting and HMDB-51 with 40$\%$ labeled setting.
	}
\resizebox{1.0\linewidth}{!}{
		\setlength{\tabcolsep}{0.5em}%
\begin{tabular}{ccccccc}
\toprule
\multirow{2}{*}{Baseline} & \multirow{2}{*}{ACL} & \multirow{2}{*}{MTL} & \multicolumn{2}{c}{UCF101-1$\%$} & \multicolumn{2}{c}{HMDB51-40$\%$} \\ \cline{4-7} 
                          &                      &                      & Top-1          & Top-5         & Top-1          & Top-5         \\ \hline
$\checkmark$ & $\times$     & $\times$     & 48.74          & 72.59          & 62.48          & 88.54          \\
$\checkmark$ & $\checkmark$ & $\times$     & 57.78          & 79.57          & 64.12          & 90.52          \\
$\checkmark$ & $\times$     & $\checkmark$ & 53.34          & 73.94          & 63.53          & 88.82          \\
$\checkmark$ & $\checkmark$ & $\checkmark$ & \textbf{60.14}          & \textbf{82.18}          & \textbf{64.90}          & \textbf{90.98}          \\ \bottomrule
\end{tabular}
}
	\label{tab:ablation-on-components}
\end{table}

%% file: Figure/abl_acl1.tex
\begin{table}[]
\centering
\caption{
		~\textbf{Analysis of Adaptive Contrastive Learning.} We compare different contrast learning strategies. The results are reported on UCF-101 with 1$\%$ labeling ratio.
	}
\resizebox{0.95\linewidth}{!}{
		\setlength{\tabcolsep}{0.5em}%
\begin{tabular}{cccccc}
\toprule
\multirow{2}{*}{Method}    & \multirow{2}{*}{Batch Size} & \multirow{2}{*}{Epochs} & \multicolumn{2}{c}{UCF101-1$\%$} \\ \cline{4-5} 
                           &                             &                         & Top-1         & Top-5         \\ \hline
Baseline                     & 5                           & 30                      &  48.74             & 72.59              \\
SimCLR~\cite{chen2020simple}                     & 5                           & 30                      &  56.25             & 79.25              \\
MoCo~\cite{he2020momentum}                       & 5                           & 30                      &  58.13             &  80.39             \\
ACL & 5                           & 30                      &   \textbf{60.14}           &   \textbf{82.18}            \\ 
\bottomrule 
\end{tabular}
}
	\label{tab:acl1}
\end{table}

%% file: Figure/abl_acl2.tex
\begin{table}[]
\centering
\caption{
		~\textbf{Analysis of threshold $\varepsilon$.} We study the effect of the hyperparameters $\varepsilon$. The results are reported on UCF-101 with 1$\%$ labeling ratio.
	}
\resizebox{0.9\linewidth}{!}{
		\setlength{\tabcolsep}{0.5em}%
\begin{tabular}{cccc}
\toprule
\multirow{2}{*}{Threshold $\varepsilon$} & \multirow{2}{*}{\begin{tabular}[c]{@{}c@{}}whether consider \\ the confidence of $x^{u}$\end{tabular}} & \multicolumn{2}{c}{UCF101-1$\%$} \\ \cline{3-4} 
                             &                                              & Top-1           & Top-5          \\ \hline
0.3                          &       $\checkmark$                                       & 58.21           & 80.46          \\
0.5                          &     $\checkmark$                                         & 59.22           & 81.53          \\
0.7                          &    $\checkmark$                                          & \textbf{60.14}           & \textbf{82.18}          \\
0.9                          &     $\checkmark$                                         & 58.62           & 80.52          \\
0.7                          &     $\times$                                         & 59.13           & 80.70          \\ \bottomrule
\end{tabular}
}
	\label{tab:acl2}
\end{table}

%% file: Figure/abl_mtl.tex

\begin{table}[t]
\centering
\caption{
		~\textbf{Analysis of Multi-scale Temporal Learning.} We study the impact of temporal information at different scales. The results are reported on UCF-101 with 1$\%$ labeling ratio.
	}
\resizebox{0.7\linewidth}{!}{
		\setlength{\tabcolsep}{0.5em}%
\begin{tabular}{cccc}
\toprule
\multirow{2}{*}{} & \multirow{2}{*}{Long-term clips} & \multicolumn{2}{c}{UCF101-1\%} \\ \cline{3-4} 
                  &                                  & Top-1          & Top-5         \\ \hline
(1)               &  None                                &  57.78              &  79.57             \\
(2)               &   Long-term $x^{L_{1}}$                               &   59.11             &  81.15             \\
(3)               &  Long-term $x^{L_{2}}$                                &  59.24              &   81.73            \\
(4)               &  MTL                                &  \textbf{60.14}              &   \textbf{82.18 }           \\ \bottomrule
\end{tabular}
}
	\label{mtl}
\end{table}

%% file: Figure/tsne.tex
\begin{figure}[t]
    \centering
    \begin{subfigure}{0.47\linewidth}
        \centering 
        \includegraphics[width=1.0\linewidth]{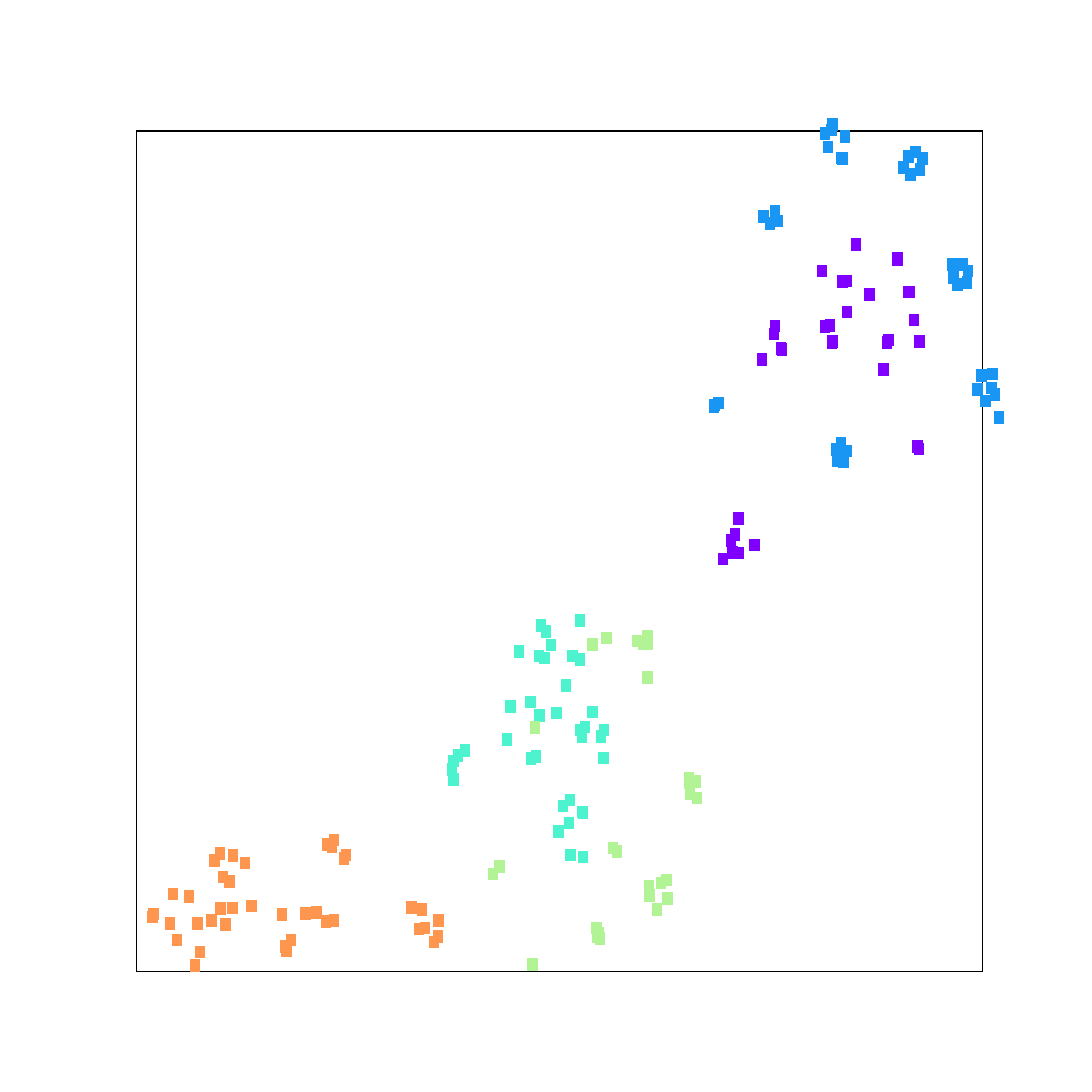}
        \caption{Base t-SNE for 5 classes}
    \end{subfigure}
    \centering
    \begin{subfigure}{0.47\linewidth}
        \centering 
        \includegraphics[width=1.0\linewidth]{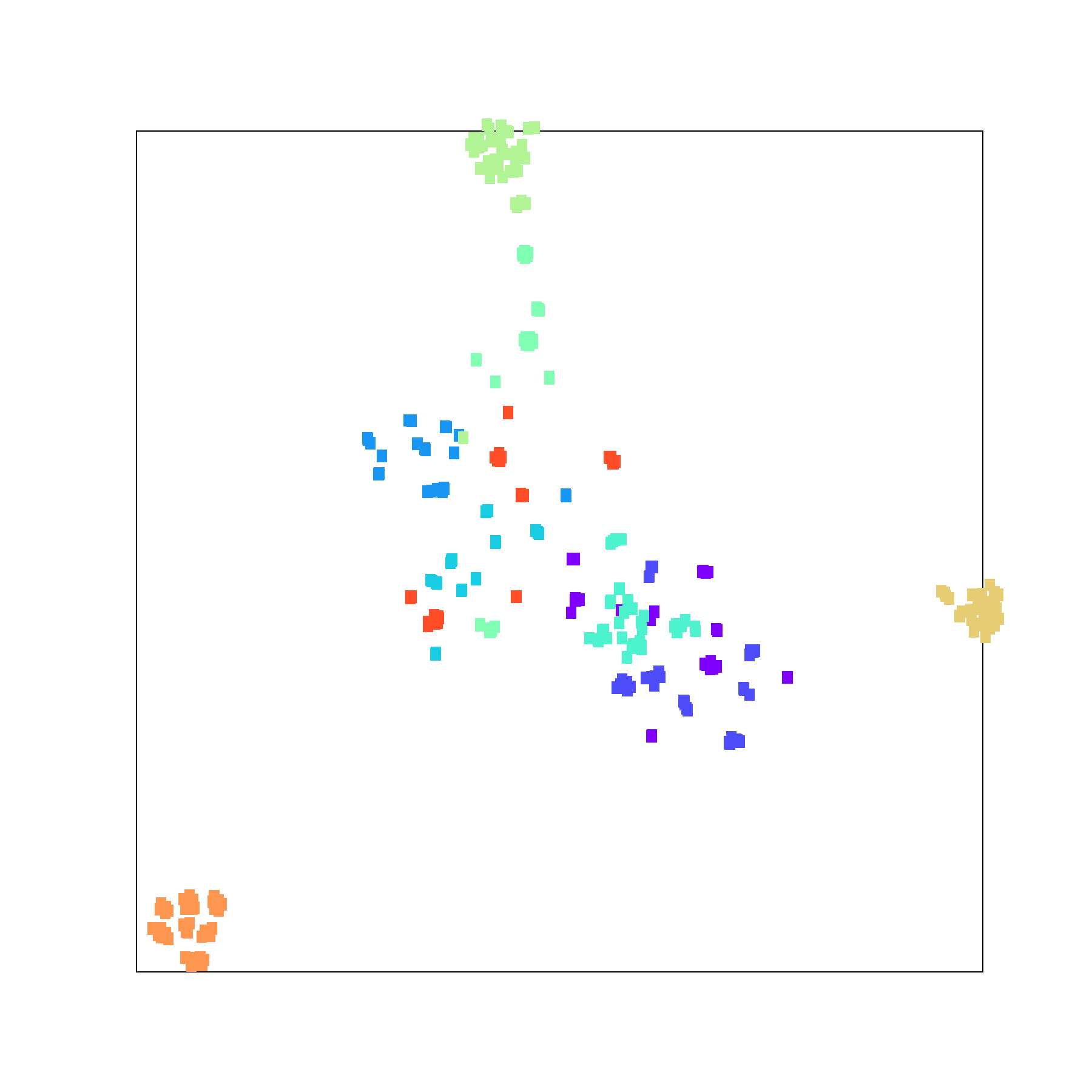}
        \caption{Base t-SNE for 10 classes}
    \end{subfigure}
    
    \centering
    \begin{subfigure}{0.47\linewidth}
        \centering 
        \includegraphics[width=1.0\linewidth]{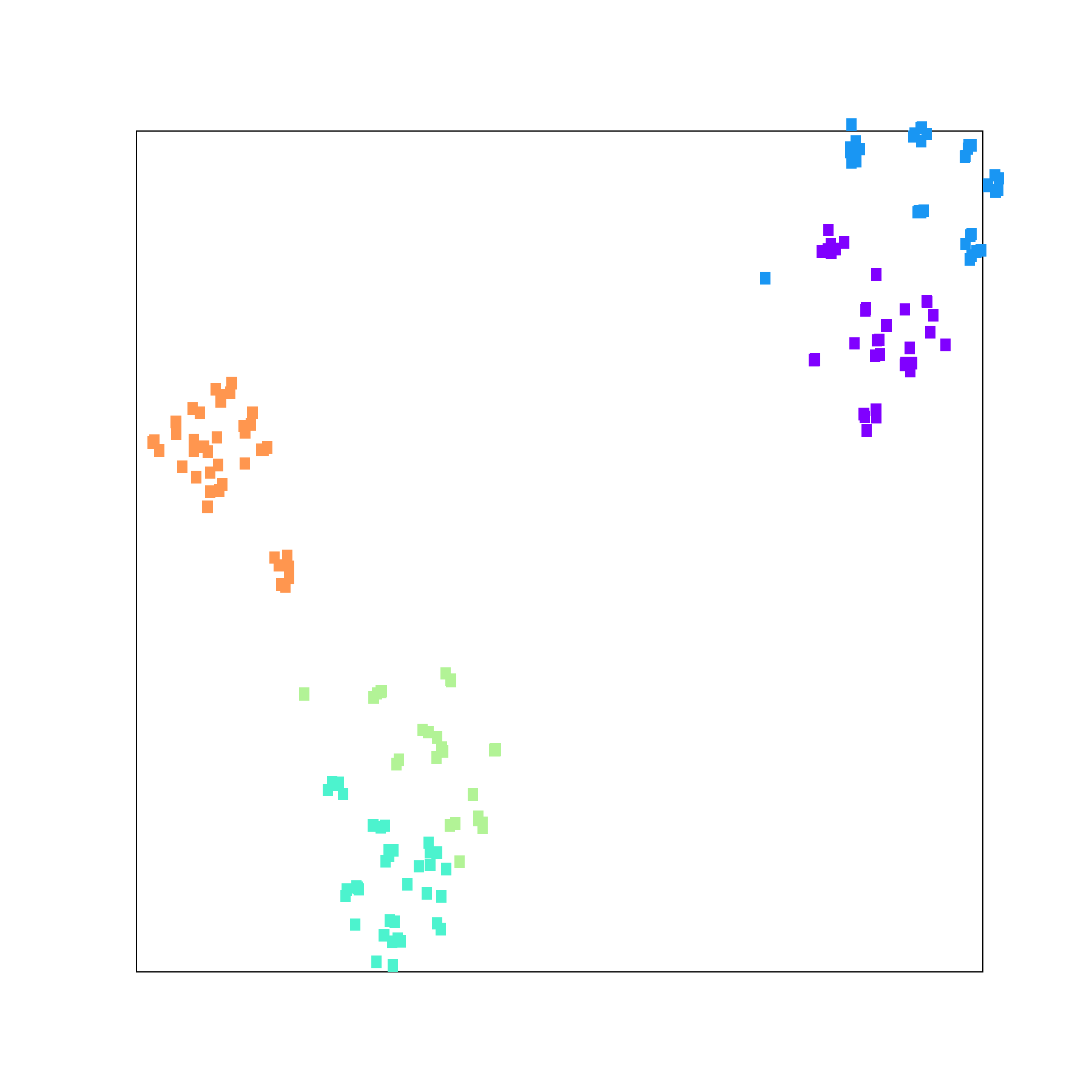}
        \caption{Our t-SNE for 5 classes}
    \end{subfigure}
    \centering
    \begin{subfigure}{0.47\linewidth}
        \centering 
        \includegraphics[width=1.0\linewidth]{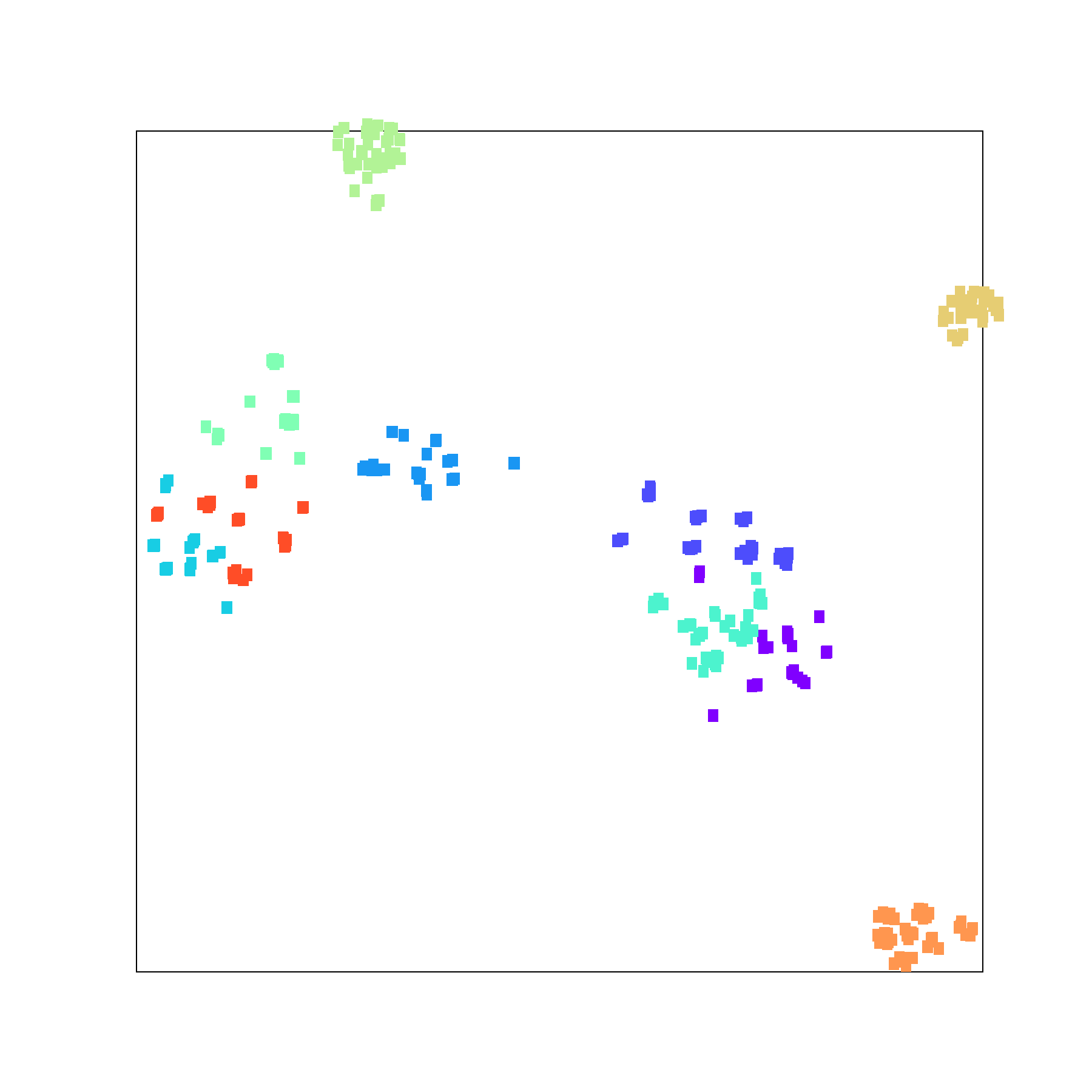}
        \caption{Our t-SNE for 10 classes}
    \end{subfigure}
    \caption{t-SNE of features on UCF-101 dataset with 1$\%$ labeled setting. The top row shows \textbf{base} features for 5 and 10 categories respectively whereas the bottom row shows \textbf{our} features after learning discriminative spatio-temporal representations. Dots of different colors represent different classes.}
    \label{fig:tsne feature}
\end{figure}

%% file: Figure/pred.tex
\begin{figure}[t]
    \centering
    \begin{subfigure}{0.47\linewidth}
        \centering 
        \includegraphics[width=1.0\linewidth]{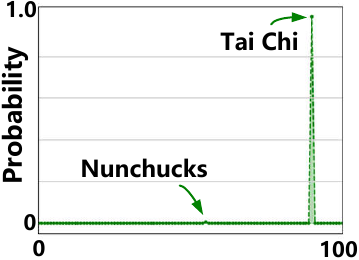}
        \caption{Base for "Nunchucks"}
    \end{subfigure}
    \centering
    \begin{subfigure}{0.47\linewidth}
        \centering 
        \includegraphics[width=1.0\linewidth]{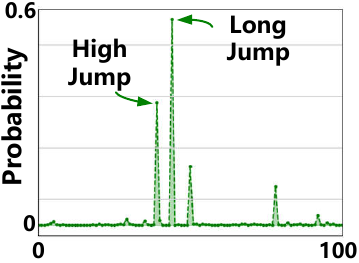}
        \caption{Base for "High Jump"}
    \end{subfigure}
    
    \centering
    \begin{subfigure}{0.47\linewidth}
        \centering 
        \includegraphics[width=1.0\linewidth]{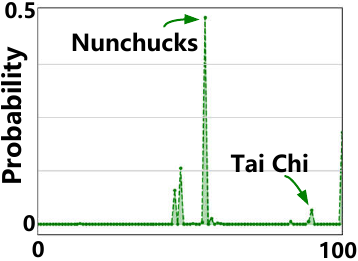}
        \caption{Ours for "Nunchucks"}
    \end{subfigure}
    \centering
    \begin{subfigure}{0.47\linewidth}
        \centering 
        \includegraphics[width=1.0\linewidth]{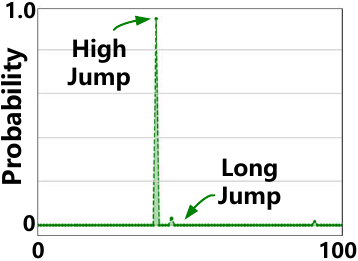}
        \caption{Ours for "High Jump"}
    \end{subfigure}
    \caption{Predictions for "Nunchucks" and "High Jump" on the UCF-101 dataset with 20$\%$ labeled setting. The top row shows the \textbf{base} predictions for the actions "Nunchucks" and "High Jump," respectively, while the bottom row presents \textbf{our} predictions after learning discriminative spatio-temporal representations.}
    \label{fig:pred}
\end{figure}

%% file: Figure/gain.tex

\begin{table}[]
\centering
\caption{Major performance gains obtained by our model over SVFormer\cite{xing2023svformer} on test categories.}
\begin{tabular}{ccc}
\toprule
\multicolumn{1}{c}{action class} & \multicolumn{1}{c}{SVFormer} & \multicolumn{1}{c}{Ours} \\ \midrule
Javelin Throw            & 0.06                       & 0.42\textbf{(+0.36)}     \\ 
Soccer Juggling         & 0.00                      & 0.49\textbf{(+0.49)}              \\ 
High Jump              & 0.03                      & 0.22\textbf{(+0.19)}              \\ 
Juggling Balls                      & 0.00                      & 0.50\textbf{(+0.50)}               \\ 
YoYo             & 0.08                     & 0.53\textbf{(+0.45)}              \\ \bottomrule
\end{tabular}
	\label{tab:performance}
\end{table}